\documentclass[12pt]{article}
\usepackage{amsmath,amsfonts,amssymb}   
\usepackage{graphicx}
\usepackage{fancyvrb}
\usepackage{color}
\usepackage{url}
\usepackage{svg}
\usepackage{hyperref}
\usepackage{caption}
\usepackage{authblk}
\usepackage{geometry}
\usepackage{natbib}
\title{A Drug Recommendation System (Dr.S) for cancer cell lines}
\author[1,2,5]{Marleen Balvert}
\author[3,5]{Georgios Patoulidis}
\author[3,5]{Andrew Patti}
\author[4]{Timo M. Deist}
\author[3]{Christine Eyler}
\author[2]{Bas E. Dutilh}
\author[1,2]{Alexander Sch\"{o}nhuth}
\author[3]{David Craft}
\affil[1]{Life Sciences \& Health, Centrum Wiskunde \& Informatica, Amsterdam, 1098 XG, The Netherlands}
\affil[2]{Theoretical Biology \& Bioinformatics, Utrecht University, Utrecht, 3512 JE, The Netherlands}
\affil[3]{Massachusetts General Hospital and Harvard Medical School, Department of Radiation Oncology, Boston MA}
\affil[4]{The D-Lab, Dpt of Precision Medicine, GROW - School for Oncology and Developmental Biology, Maastricht University Medical Centre+, Maastricht, The Netherlands}
\affil[5]{Equal contribution}
\date{December 24, 2019}

\begin{document}

\maketitle
\thispagestyle{empty}

\begin{abstract}
    Personalizing drug prescriptions in cancer care based on genomic information requires associating genomic markers with treatment effects. This is an unsolved challenge requiring genomic patient data in yet unavailable volumes as well as appropriate quantitative methods. We attempt to solve this challenge for an experimental proxy for which sufficient data is available: 42 drugs tested on 1018 cancer cell lines. 
    Our goal is to develop a method to identify the drug that is most promising based on a cell line's genomic information. For this, we need to identify for each drug the machine learning method, choice of hyperparameters and genomic features for optimal predictive performance.
    We extensively compare 
    combinations of gene sets (both curated and random), genetic features, and machine learning algorithms for all 42 drugs. For each drug, the best performing combination (considering only the curated gene sets) 
    is selected. We use these top model parameters for each drug to build and demonstrate a Drug Recommendation System (Dr.S). Insights resulting from this analysis are formulated as best practices for developing drug recommendation systems. The complete software system, called the Cell Line Analyzer, is written in Python and available on github.
    \end{abstract}
    
\section{Introduction and motivation}


Personalized genomic medicine for the majority of cancer patients is still not a reality.  Although some patients have genetic testing done on tumor biopsies which occasionally leads to identification of an actionable mutation, genetic screening is far from routine clinical practice. Even when patients are screened, it is still largely unknown which genetic characteristics are indicative of a drug being effective. Actionable genetic features are mainly single nucleotide polymorphisms (SNPs) of known cancer-related genes, and less frequently chromosomal translocations \citep{personalizedCancer1}. Cell-line drug screens show that drugs designed and approved for specific tissue types can be potent in other tissues as well \citep{yang2012genomics}. This makes drug prescription based on genomic characteristics a promising approach to cancer treatment.

Since the creation of large cell line-drug sensitivity datasets such as the Genomics of Drug Sensitivity in Cancer (GDSC, \citep{yang2018linking}) and the Cancer Cell Line Encyclopedia (CCLE, \citep{barretina2012cancer}), many researchers have attempted to build machine learning models that predict drug sensitivity from genomic properties of the cell lines, reviewed in \cite{ali2019machine}. A wide variety of approaches has been used, varying not just in terms of the machine learning algorithms, but also in the type of input data that is included in the analysis, the usage of data in the experimental design (data splitting, e.g. whether to use training and validation data only or use test data as well), the measure used for performance assessment, and more. We provide an extensive literature review in the Supplementary Information (SI). To the best of our knowledge, all of the research on machine learning for these datasets builds models to predict drug sensitivity. The next step, creating a system that can recommend a (combination of) drug(s) for a new cell line (as a prototype personalized drug recommendation system for patients), has to date not been explored.

To build a clinical patient drug recommendation system would require databases with clinical variables, genomic information, and well-curated per-drug outcomes for tens of thousands of patients. As such databases are not available we instead work exclusively with cell line datasets, \emph{but do so in a way that mimics how a clinical drug recommendation system would be built}. 
In doing so, we promote several best practices for working with such datasets, including proper data hold out strategies \citep{ESL} and the inclusion of prior knowledge \citep{priorknowledge} through {\em a~priori} gene set selection. We demonstrate that 1) gene expression is, by far, the most important genomic variable type for predicting a cell line's response to a drug (amongst gene expression, mutation, and copy number), 
2) prior knowledge gene sets outperform both gene sets based on univariate correlation tests and random gene sets of the same number of genes, although random gene sets that are larger than our largest curated set occasionally outperform the curated sets, and 3) although the coefficient of determination ($R^2$) values of learning cell line sensitivity to individual drugs is on average low (averaging around 0.3, which is consistent across the literature), our drug recommendation system (Dr.S) is nevertheless able to recommend drugs that are (among) the best drugs for the considered cell line and outperforms a tissue type based drug recommendation.

We built a software system called the Cell Line Analyzer (CLA) which consists of two modules. The first module, Model Analysis and Selection (MAS), is used to explore how ``learnable'' each drug is. For each drug it cycles through user-selected machine learning algorithms, their tunable hyperparameters, and combinations of gene sets and genetic feature types (e.g. expression, mutation, and copy number), to determine which choices lead to the highest quality predictive models. 

The second module, the Drug Recommendation System (Dr.S), uses the same datasets but has a more practical (and clinically inspired) goal in mind: for a not-yet-seen cell line (i.e. a held out cell line), we aim to recommend a drug that will most likely destroy the cell line. To maximize performance in this module, we retrain models for each held out cell line using the optimal parameters (drug specific) found in the MAS explorations. This is a first step towards building a drug recommendation system for a clinical setting and lays out the basic concepts of such a system. 


\section{Methods}

\subsection{Data}
We use cell line data from the GDSC \citep{garnett2012systematic, yang2012genomics} version 17.3, which comprises gene expression, mutation and copy number variation information of 1018 cell lines. At the time that we processed the data (March 2018), dose response curves of these cell lines with respect to 250 unique drugs were available. We only consider 42 drugs for which we were able to obtain reasonable drug sensitivity predictions in an earlier study ($R^2\geq 0.2$ \cite{georgeThesis}). In the original drug sensitivity wetlab measurement experiments performed by GDSC, each drug was applied to a subset of these cell lines (median number of cell lines that a given drug was applied to: 892). For Dr.S we only included cell lines that were tested for at least 15 drugs, which is a total of 943 cell lines.

In this work we do not only include drugs, but look at radiation therapy as well. Radiation sensitivity information is obtained for the CCLE data from \cite{yard2016genetic}, and this dataset likewise contains gene expression, mutation and copy number variation data of 524 cell lines. The radiation data is not used in the Dr.S analysis, since it uses a different set of cell lines.

\subsection{Cell Line Analyzer, Module 1: Model Analysis and Selection (MAS)}

For each drug of interest our basic task is to build a model which, given a new cell line, predicts the potency of that drug on that cell line. All of the data handling decisions we make are based on the guiding motivation of: ``How would we create this tool if we were building a drug recommendation system with real patient data for use on future patients?'' The goal of the exploratory MAS phase is twofold: we will investigate the ``learnability'' of the relationship between genomics and drug sensitivity, and for each drug we will select the models (machine learning algorithms, hyperparameters, and the combination of gene sets and feature types) that give the best predictive performance to be used in Module 2, Dr.S.

In order to leverage prior knowledge in building models that predict drug response, we utilize curated gene sets. The six sets that we use (five of them from the literature, one of them constructed by our group from biological principles, without looking at the data, see SI section 4) have between 65 and 263 genes, 148 genes on average. Some genes are in more than one of the six sets, but the sets have minimal overlap (see SI Table 8). We use three genomic feature types: gene expression, mutation, and copy number. A ``combo'' is defined as a gene set (or possibly ``none'') for each of expression, mutation, and copy number, e.g. expression: gene set 1, mutation: gene set 6, copy number: gene set 4. For each drug we assess all possible combos ($7^3 - 1$ = 342) with a repeated holdout procedure (SI sections 1 and 2). 

We tune and train three types of machine learning algorithms: elastic net, support vector machine regression with radial basis function kernel, and random forest. These algorithms represent a breadth of high quality off-the-shelf machine learning approaches suitable for datasets with a limited number of samples (cell lines). We use a strict data holdout strategy to assess how well machine learning models will do on never-before-seen data (SI section 2), and we perform multiple holdouts (independent random sampling) in order to get statistics on the predictive algorithms. As we will show, there is substantial variation of model performance over different holdout sets, and reporting results for a single holdout set, which research groups often do, will not be representative. The complete description of the gene set combos, data holdout, hyperparameter tuning, and model assessment procedures is given in the SI. Note that our strict data splitting policy prohibits an initial dimension reduction step, such as principle components analysis or univariate feature selection, on the entire dataset. The correct approach is to perform the dimension reduction step with the training data only, that is, without looking at the holdout data, discussed below. 

Our measure of model quality is the coefficient of determination, $R^2$, on the holdout data. Consider a particular drug. Let $y_i$ be the actual response of cell line $i$ to that drug. In MAS, we use the area under the dose response curve (AUC) as $y_i$, which is a continuous value on $[0,1]$. Let $\hat y_i$ be the model predicted response and let $\bar y$ be the mean value of the true responses in the holdout data. We then compute:
\begin{equation}
    R^2 = 1 - \frac{\sum_i (y_i - \hat y_i)^2}{\sum_i (y_i - \bar y)^2}
\end{equation}
\noindent where the sums are over the holdout samples. If our predictions are perfect $R^2 = 1$, and if we always predict the holdout mean, $R^2=0$. Note that with this definition, it is possible to have negative $R^2$, which simply means the predictions are overall worse than predicting the mean.

We work with each drug individually rather than attempt a multi-drug combined learning model. The reason for this is that we are simulating the clinical context, where the training data will consist of patients and the treatment(s) they were given. For predicting drug sensitivity of a cell line for a certain drug $d$ a multi-drug learning model uses drug sensitivity information of that cell line for a different set of drugs as a characteristic feature of the cell line, searches for cell lines with similar features (i.e. cell lines that responded similarly to those drugs) and uses the drug sensitivity of those cell lines for drug $d$ to predict sensitivity of the current cell line to drug $d$. Unlike the cell line databases, where every cell line can be given every drug, patients are given a comparatively small number of drugs. When clinically building up
models of efficacy and toxicity for a particular drug, one will identify a large number of patients given that drug, and this group of patients will be different than the cohort used to build the model for a different drug. Thus we consider each drug independently.

Feature importances are computed for elastic net and random forest. For elastic net, we use the normalized coefficients of the regression as feature importances. For random forest, the python package sklearn computes them using the mean purity decrease at nodes that split on a given feature \citep{sklearn}. 

\subsubsection{Univariate-based feature selection}
We assess a statistically valid approach to univariate feature selection, where the features are selected after data is held out (that is, without having access to the holdout data). This contrasts the common approach where variables are selected based on the entire dataset. The idea of univariate selection is to choose features that correlate well on their own with the outcome. We chose to compare our biological knowledge based gene selection approach against univariate feature selection since it is a common feature selection approach which is not computationally demanding due to its linear nature, a requirement for a dataset with a large number of features such as the GDSC. To compare univariate feature selection to the prior knowledge gene set technique that constitutes our main approach, since the 
maximum gene set size is 263 genes we use univariate selection to select 263 genes for each of expression, copy number, and mutation. 
Expression and copy number are quantitative variables so for each we select 263 genes based on the top Spearman correlations. 
For mutation data, we first reduce the seven categories to two (mutated or not; we verified that this collapse does not alter the baseline results, see SI section 8) and then use the rank-sum test for the univariate selection. 

\subsubsection{Curated versus randomly selected genes}
For all of our Dr.S runs--described in the next section--we use our six curated gene sets. However, 
to assess the added value of currently available prior knowledge compared to a random selection of genes, we also use the MAS module to study these sets versus sets of various sizes consisting of randomly selected genes. 
We compare the performance of our curated sets as well as the union of these six curated sets (representing prior knowledge) versus various sets of randomly selected genes consisting of between 125 and 1,500 genes.

\subsection{Cell Line Analyzer, Module 2: Drug Recommendation System (Dr.S)}
For Dr.S we assume that each drug will be recommended at a particular dose. We use the normalized concentration levels provided by the GDSC, which range from 0 to 9 \citep{vis}. The drug response of a cell line at a certain concentration level is measured in terms of viability: the fraction of cell lines that survived for 72 hours after applying the drug at the chosen concentration level. For each drug, we choose a concentration level such that the average viability across the cell lines hit by that drug is 0.75. We use this as a surrogate for normal tissue toxicity levels: if the viability of a drug at a certain concentration level averaged over all cell lines is low, we assume the drug to be toxic and vice versa. By choosing a concentration level that yields the same average viability for all drugs we implicitly assume that each drug is equally toxic at its individual concentration level.

In Dr.S, we use a leave-one-out strategy to push the limits of our model quality. For each cell line, which represents a new incoming ``patient'', we consider the set of drugs, from the 42 drugs we study, that were applied to that cell line experimentally. For each of these drugs we select the best machine learning approach (elastic net, random forest or support vector machine), the combo and the hyperparameters that are found by MAS. A model is then trained to predict viability at the drug concentration as detailed above, using all the cell lines that drug was tested on experimentally except the held out cell line. After training, the models are used to predict the viability of the held out cell line for each drug. Dr.S outputs a sorted set of drugs and their predicted viabilities for that cell line, which are then further processed into a (set of) recommendation(s). This training and recommendation procedure is repeated for every cell line to understand the overall quality of the recommendation system. 

We investigate two drug recommendation policies. The first selects the $N$ drugs with the lowest predicted viability. While simple and intuitive, having a strict count cutoff does not always make sense so we also examine the strategy where we recommend the set of drugs where the predicted viability is within some $\varepsilon$ of the lowest predicted viability.

\subsection{The complete system: the Cell Line Analyzer (CLA)}
The analysis software that contains both the MAS and Dr.S, which is called the Cell Line Analyzer (CLA), is written in Python and is available at \url{ https://github.com/aspatti1257/CellLineAnalyzer}. For an MAS run, for a particular drug, the user supplies gene sets, gene feature files (gene expression, etc), the response variable to learn, and an arguments file saying which algorithms to run and data splitting values. For a Dr.S run one uses the same set of files as for an MAS run, supplemented with the output files of the MAS which contain the selected algorithms and hyperparameters for each drug. All documentation is at the available github site.

\section{Results}

\subsection{Part 1: MAS}

\begin{figure}[!htbp]
  \centering
  \includegraphics[width=\linewidth]{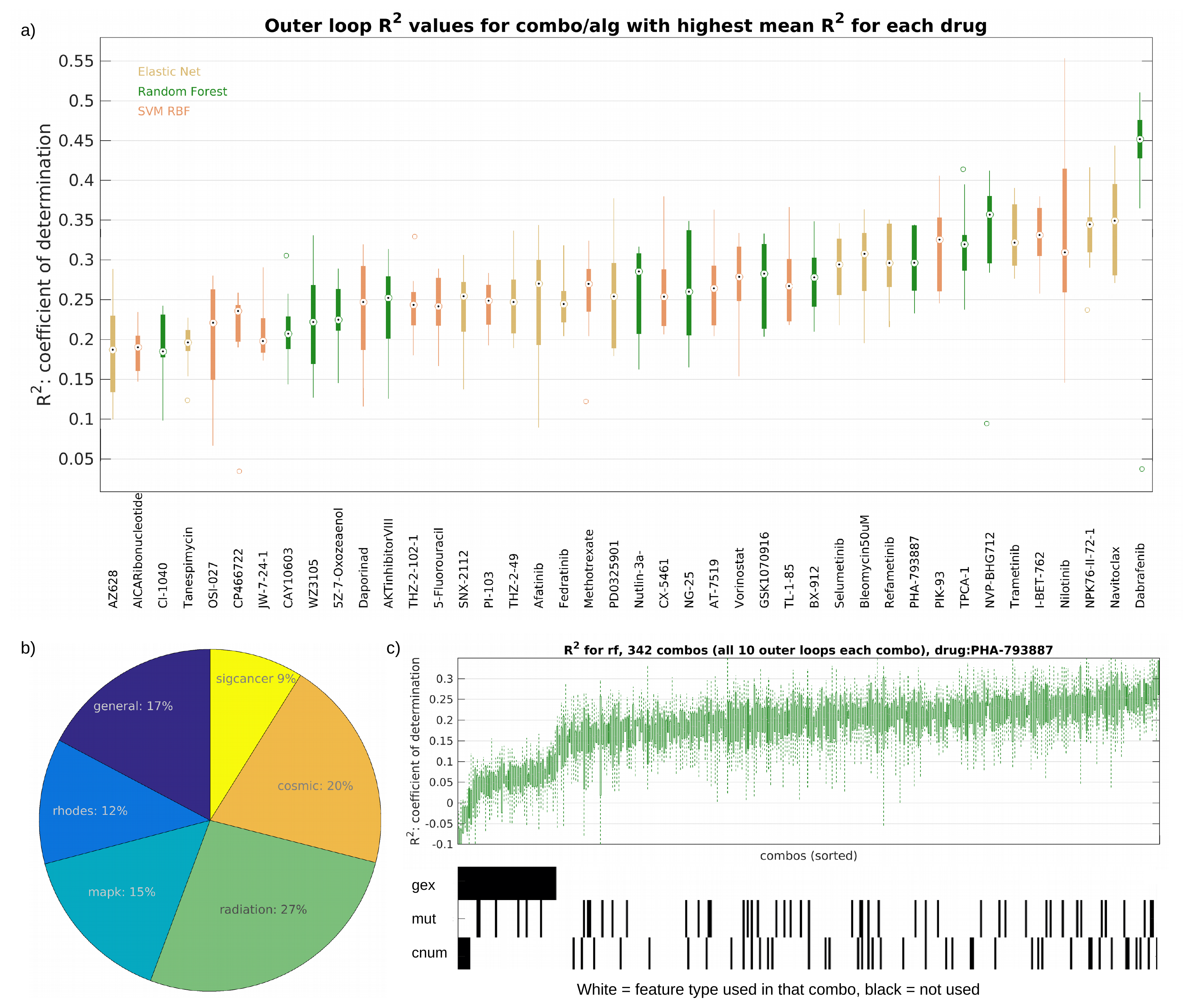}
  \caption{Summary of baseline MAS results. The upper panel shows the $R^2$ values for the top performing (top mean $R^2$ over the ten outer holdout sets) algorithm (color coded) and combo (combo specifics not shown). The drugs are sorted from left to right by increasing mean $R^2$. The lower left pie chart displays the relative usage of the six gene sets for the top five combos for each of the three ML algorithms: every time a particular set was used in a top performing combo (which could be more than once) that gene set counter was incremented. Thus, radiation is the most frequently used in top performing combos, and it is used e.g. three times as often as the sigcancer gene set. The lower right panel displays the results of all 342 combos for one drug (PHA-793887) and one algorithm (Random Forest), and for each combo all 10 outer loop $R^2$ results are box plotted. The combos are sorted by increasing mean $R^2$. The black and white indicator plot below is aligned with the individual combos and displays if a feature type was used (white) or not (black) in that combo. gex=gene expression, mut=mutation, cnum=copy number.}
  \label{DrugSummary}
\end{figure}

Figure \ref{DrugSummary} summarizes the MAS results for the 42 GDSC drugs. We see that each of the algorithm types was selected as the top performing one equally often (elastic net: 13, support vector machine: 16, random forest: 13), and that the drugs differed in predictability from median $R^2=0.19$ for AZ628 to median $R^2=0.45$ for Dabrafenib. The most frequently used gene set among the top five performing combos across all drugs was the radiation gene set, followed by the cosmic gene set. These two sets are also the largest, which is likely influential; we analyze this finding further in SI section 11.2. More detailed combo information for a single drug and algorithm is shown in Figure \ref{DrugSummary}c, which also displays that without gene expression information, the $R^2$ values dropped by over 10\%. This is true across most drugs (SI Figure 1).

\begin{figure}[!htbp]
  \centering
  \includegraphics[width=\linewidth]{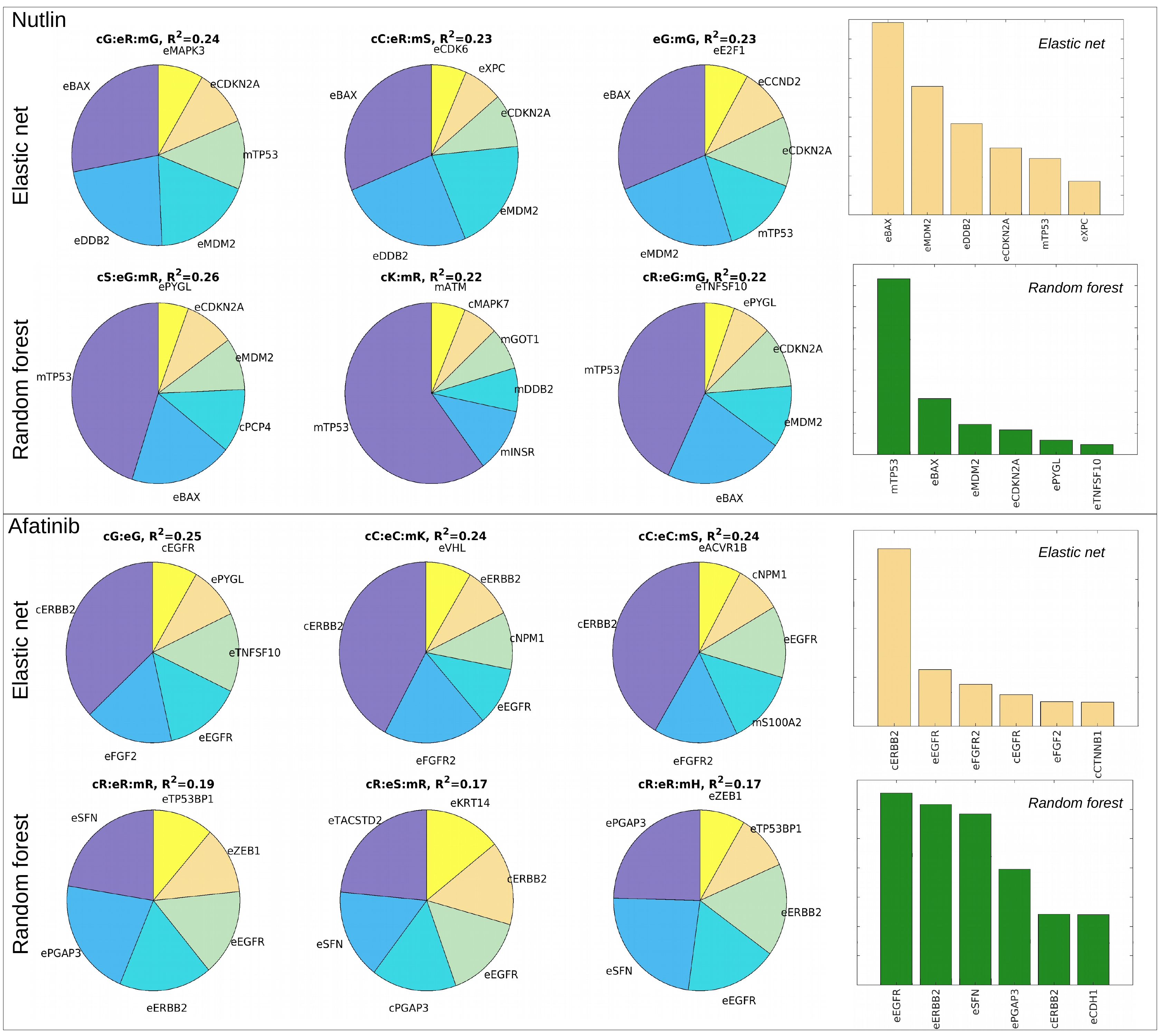}
  \caption{Feature importances for Nutlin-3a (top panel) and Afatinib (lower panel). For each drug, the top row shows results for elastic net, and the second row random forest. The combo used is written in shorthand above each pie chart. Lowercase letter indicates expression, mutation, or copy number, followed by a single uppercase letter denoting the gene set: R=radiation, S=sigcancer, G=general, H=Rhodes, C=cosmic, K=mapk. The three pie charts for each drug/algorithm are the best scoring combo (left most) and then the next two best combos proceeding rightward. The bar graphs summarize the feature importances considering the top 10 combos. Gene names are preceded by a single lowercase letter indicating expression, mutation, or copy number.}
  \label{fi}
\end{figure}

In Figure \ref{fi} we highlight feature importances (FI) for two drugs: Nutlin-3a and Afatinib. FI are available for every drug in the SI Section 6. Many drugs, including these two, confirmed our approach by displaying known drug interactions as top features. For example we see that for Nutlin-3a, our models found TP53 and BAX as strong predictors: Nutlin-3a disrupts the P53 pathway by interfering with the interaction between P53 and MDM2, and BAX is an apoptotic activator which is regulated by P53 \citep{weinberg}. Afatinib is a tyrosine kinase inhibitor and is known to interact with EGFR and ERBB2 \citep{afatinib}. In nearly all of the FI plots (SI Figure 2) we see that elastic net and random forest revealed distinct features as important, which likely relates to the fact that elastic net determines only features that are linearly related to the response variable, while random forest can capture non-linear relationships as well.

For none of the 42 drugs did the univariate selection process outperform the use of prior knowledge based gene sets, while for 43\% of the drugs (18/42),
the univariate selection process was significantly worse than the top gene set combo. These results are summarized in SI Figure 3. 
We used 0/1 mutation encoding (reflecting whether a gene is mutated or not) for the univariate runs while we used a 7-class label for the regular MAS analysis reflecting the type of mutation that occurs in a gene (missense, nonsense, frameshift, etc). We therefore independently verified that the choice of encoding does not affect the results by re-running the MAS for three of the drugs (Vorinostat, Selumetinib, and TL-1-85; randomly chosen from the mid-range of predictability) with 0/1 mutation encoding. The results were not significantly different between types of encoding, as shown in SI Figure 4. Including tissue type of the cell line as a categorical feature variable did not significantly alter the results either, see Figure SI 5.


\subsubsection{Curated gene sets compared to randomly selected genes}
In order to investigate how the size and content of the gene sets influences the prediction quality, we perform a series of runs involving randomly selected genes. For all of the runs described in this section, we restrict ourselves to mutation and gene expression data, since copy number, as judged from the feature importances sets, is the least valuable data type.

Our first set of runs compares our top curated list, the radiation list of 263 genes, with two lists of randomly selected genes, each of size 263. These three sets were then sent through the MAS for all 42 drugs. For each of the three gene sets, we examine the $R^2$ values for all the combos that use that set and all that do not use it. We compare these two result sets (with and without a gene set) with a rank sum test to see if there is a statistically significant difference when we leave out a gene set, see SI Figure 7, the left heatmap. We see that leaving out the curated radiation set is overall more detrimental: for 13 of the 42 drugs the p-value is $<0.05$, compared to only 1 drug for random gene set 1, and 5 for random gene set 2. We also examine how many times each set is used in the top performing combo for each drug, results shown in the pie graph of SI Figure 7. Thus, we conclude that the radiation gene set is superior, but only for 13 of the 42 drugs are the differences in $R^2$ values for combos with the radiation set vs. without the set statistically distinguishable. We repeat this procedure for the union of the curated sets, which consists of 754 genes, versus two random sets of this size. Results are similar, see SI Figure 8.

Next we generated four random gene sets of the following sizes: 125, 250, 500, and 1,000 genes. SI Figure 9 shows that the gene set of size 1,000 has the highest predictive power. When comparing the baseline MAS results to the top combos with these random gene sets (SI Figure 10), using a ranksum test to assess the difference in $R^2$, 
there are five drugs that show a statically significant difference at the $p=0.05$ level (THZ-2-102-1, Afatinib, Nutlin-3a, Nilotinib, and Dabrafenib), each with baseline MAS results superior to the random gene lists top combos.

Given that in the four random gene sets analysis the 1,000 genes set was the best of the four, we were curious if even larger random gene sets would be better. We ran the MAS for the union of the curated gene sets, containing 754 genes, along with three randomly generated sets of 754 genes, 1,000 genes, and 1,500 genes. Results are shown in SI Figures 11 and 12. The random gene set of size 1,500 is the most commonly used gene set in the top combos, appearing 38\% of the time, followed by the random 1,000 gene set. For Dabrafenib and Nutlin-3a the curated gene lists yield statistically significant better results than the random gene set ($p=0.05$). Not including genes that these drugs target, BRAF and P53-MDM2 (and thus indirectly, BAX) which are included in the curated union sets, is strongly detrimental. A side-by-side comparison of the baseline MAS results and the best combo from these runs is shown in SI Figure 12. Three drugs show significantly different results between the random and baseline curated MAS runs: AZ628 (random sets better), CI10-40 (random sets better), and Nilotinib (curated gene sets better). 

From these results we can draw the following conclusions. 1) For gene sets of equal size, custom curated gene sets outperform random gene sets, at least up to our largest curated set of 754 genes, which is the union of our six curated sets. 2) Larger random gene sets are more effective than smaller ones: for a run comparing random sets of sizes 125, 250, 500, and 1000 genes, the 1000 gene set was the most commonly used by the top models, and 3) our largest random list, of size 1500, is competitive with our union curated list of size 754 genes, but at a higher computational cost, and outperforms the baseline results for only two drugs.

\subsubsection{Radiation results}

The radiation dataset consists of 524 cell lines from the CCLE that were irradiated at various levels to create the dose response curves \citep{yard2016genetic}. SVM was the best machine learning algorithm for this dataset, where the top combo yielded $R^2 = 0.19$. The top combo did not use the radiation gene set, but the next two highest combos, which also both come in at $R^2 = 0.19$ (they differ in the next decimal places), both used the radiation gene set. The most commonly used gene set for the radiation dataset for the top five combos for each of the three algorithms was the Rhodes set. Contrary to the implications of the authors of the original radiation dataset, we found that gene expression is more indispensable than copy number for obtaining high $R^2$ values, see SI Figure 6.

The feature that came out as most important for the radiation runs was the expression level of SMARCA4 (SI Figure 2, last two rows), a transcription regulator of the SWI/SNF protein family, which is involved with the dynamic packaging and accessibility of DNA \citep{swisnf}.

\subsection{Part 2: Dr.S}

\subsubsection{Prescribing the \emph{N} drugs with lowest viability}
\begin{figure}[!htbp]
    \centering
    \includegraphics[width=\linewidth]{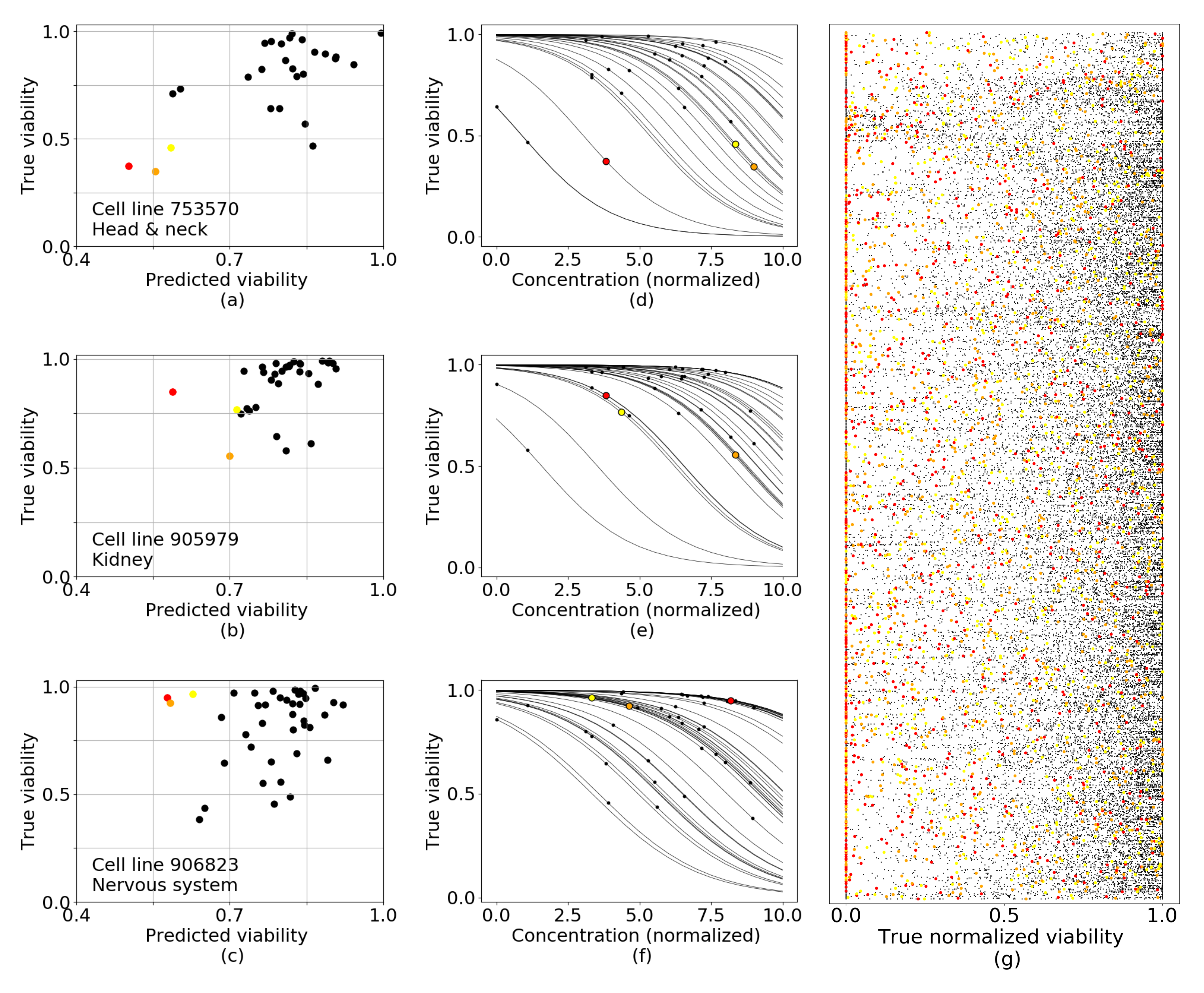}
    \caption{Overview of the performance of Dr.S. when recommending the top $N$ drugs. Figures (a)-(c) show the predicted and true viability of each drug for a cell line for which the recommendation was very good (a), mediocre (b) and very bad (c). Viabilities for the first, second and third drugs recommended by Dr.S are indicated in red, orange and yellow, respectively, viabilities for the other drugs are indicated in black. Figures (d)-(f) show the dose-response curves for the same cell lines as in Figures (a)-(c). The dots indicate the concentration at which the drug was administered and the corresponding viability, where colors are as before. In Figure (g) each dot corresponds to a cell line-drug combination. Each row (on the vertical axis) corresponds to a cell line, and the horizontal axis represents the normalized viability of that cell line with respect to a drug. A normalized viability of 0 (1) corresponds to the viability of the best (worst) drug. As before, the red, orange and yellow dots indicate the first, second and third recommended drug. The plot shows that for many cell lines Dr.S recommends a drug with low viability, which can be concluded from the colored dots being mostly on the left side of the plot.}
    \label{fig:DrS_resultfig1}
\end{figure}

Figure \ref{fig:DrS_resultfig1} shows the quality of the recommendation made by Dr.S for three individual cell lines, chosen to represent the range of prediction quality. Each dot represents a drug, and the red, orange, and yellow indicate the first, second, and third recommended drug. Dr.S performed well for cell line 753570 (top), not so well for cell line 905979 (middle) and poorly for cell line 906823 (bottom). While Figures \ref{fig:DrS_resultfig1}a-f show detailed results for three cell lines, Figure \ref{fig:DrS_resultfig1}g summarizes results for all cell lines. Each dot in the plot represents the true normalized viability (horizontal axis) of a cell line with respect to a drug at the prescribed concentration, where each row on the vertical axis corresponds to a cell line. Again, the red, orange and yellow dots represent the first, second and third recommended drug by Dr.S. As can be seen from the density of colored dots at the left in Figure \ref{fig:DrS_resultfig1}g, Dr.S. generally selected drugs for which the viability is low.

\paragraph{Dr.S outperforms a tissue type based approach}

\begin{figure}[!htbp]
    \centering
    \includegraphics[width=0.9\textwidth]{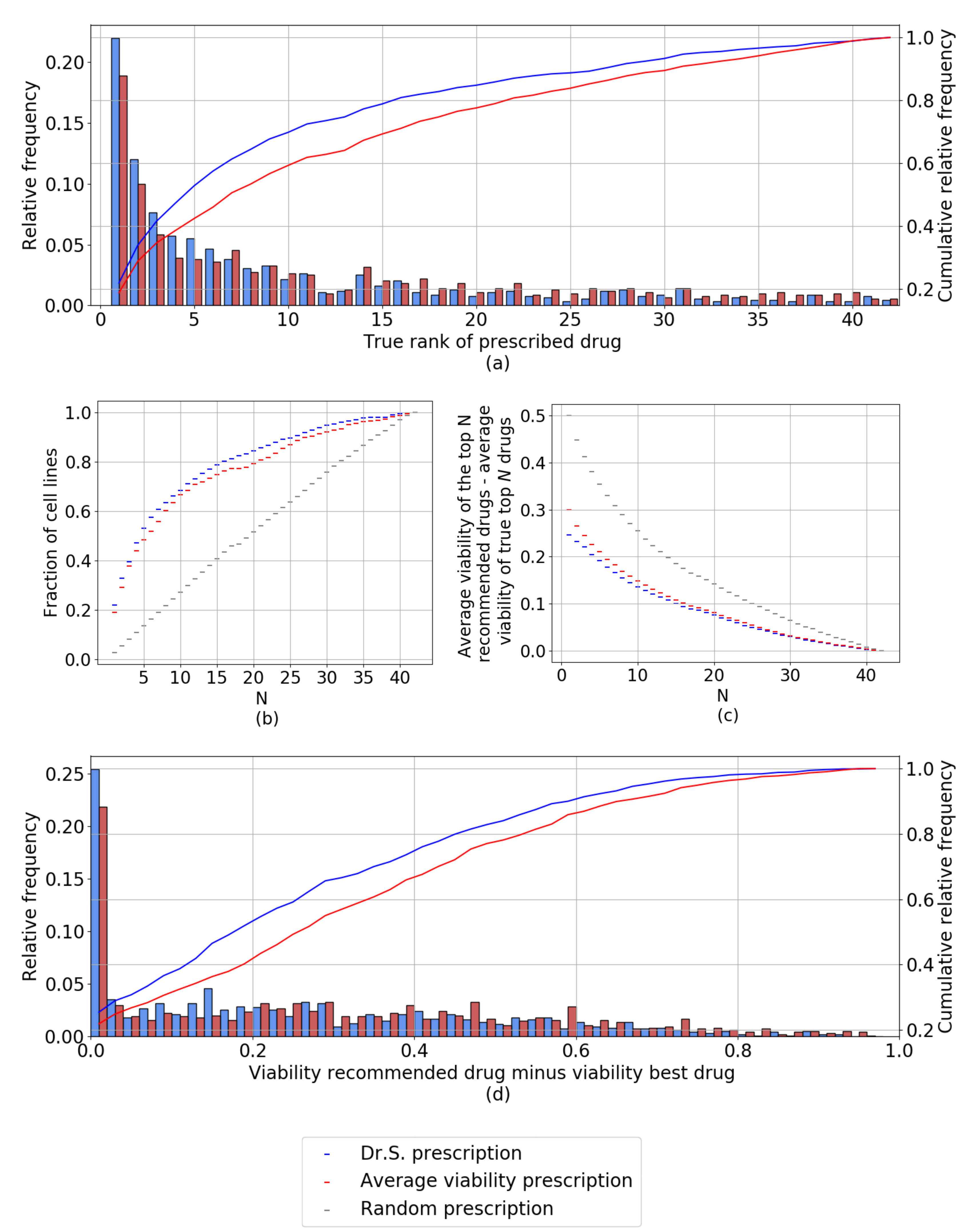}
    \caption{Comparison of drug recommendations made with Dr.S. (blue), recommendations based on the average viability of other cell lines of the same tissue type from the training data (red) and random drug selection (gray). (a) The (cumulative) distribution of the true rank of the single prescribed drug. (b) The fraction of cell lines for which the single true best drug is among the $N$ true recommended drugs. (c) The difference in viability between the top $N$ prescribed and the top $N$ true best drugs. (d) The distribution of the difference in viability between the single prescribed and single best drug.}
    \label{fig:DrS_resultfig2}
\end{figure}

While Figure \ref{fig:DrS_resultfig1} gives a visual overview of the performance of Dr.S for individual cell lines, Figure \ref{fig:DrS_resultfig2} provides summary statistics. We compared the results obtained with Dr.S. with two other policies. In the first policy, as a null policy, we randomly selected a drug. The second policy is designed to reflect a more clinically realistic situation. A sensible policy for drug recommendation would be based on past experience with similar patients (or cell lines). Therefore, when prescribing a drug to a new cell line, we considered all other cell lines of the same tissue type (see Table 9 in the SI for a description of the tissue type classification we used), where the past experience is the (true) viability of a cell line for each drug that was observed after administering a drug to that cell line. The drug that had the best average viability over all cell lines with the same tissue type as the current cell line was prescribed to the current cell line.

First we look at the case where $N=1$, i.e., we recommend only a single drug. As can be seen from Figure \ref{fig:DrS_resultfig2}a, which shows a histogram along with a cumulative distribution of the true rank of the single recommended drug over all cell lines, Dr.S recommended the true best drug for 22.0\% of the cell lines compared to 18.8\% with a tissue type based recommendation. With Dr.S the recommended drug is among the true top 5 for 52.9\% of the cell lines, while this is the case for only 42.5\% of the cell lines when recommending drugs based on tissue type. Figure \ref{fig:DrS_resultfig2}b shows the fraction of cell lines for which the true single best drug was among the recommended top $N$ drugs, a number which goes to 1 as $N$ goes to the total number of drugs. The recommendation by Dr.S. (blue) was compared to a tissue type based recommendation (red) and a randomly selected drug (gray). The fraction of cell lines for which the best drug was within the recommended set of size $N$ increases rapidly: when $N = 3$ already 39.4\% (37.8\%) of the cell lines got a Dr.S (tissue type based) recommendation that included the true best drug, and 75\% of the cell lines had the true best drug within the recommended set when $N=13$ with Dr.S and $N=16$ with a tissue type based recommendation. 

Not recommending the true best drug is not problematic as long as the viability of the recommended drug is close to the viability of the true best drug. Figure \ref{fig:DrS_resultfig2}d, a histogram of the difference between the viability of the single recommended drug and the viability of the true best drug, shows that for 25.4\% of the cell lines the viability of the recommended drug was at most 0.02 higher than the true best viability when using Dr.S, while with a tissue type based prescription 21.9\% of the cell lines were prescribed a drug with a viability within 0.02 from the true best viability. Dr.S outperformed a tissue type based prescription for almost all tissue types (20 out of 25, see SI Figure 13). Figure \ref{fig:DrS_resultfig2}c shows the difference between the average viability of the recommended $N$ drugs and the average viability of the true best $N$ drugs. When $N=1$ this was 0.247, an improvement of 17\% compared to a tissue type based prescription (0.299). 

\subsubsection{Prescribing all drugs that have a predicted viability within $\varepsilon$ of the best predicted viability}
\begin{figure}[!htbp]
    \centering
    \includegraphics[width=0.45\textwidth]{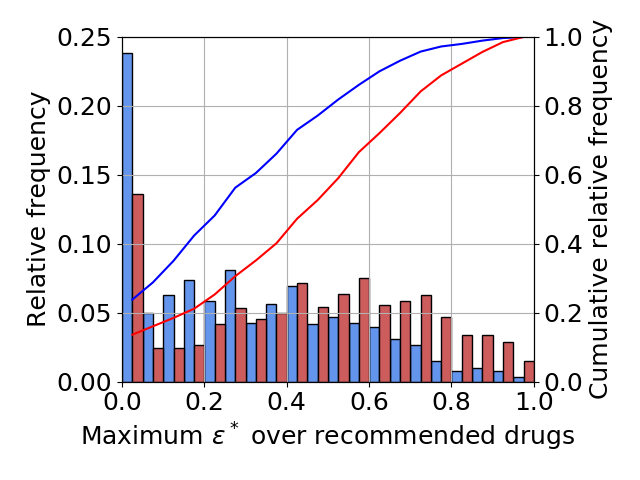}
    \caption{Results obtained with Dr.S and tissue type based prescription where all drugs with a predicted viability within $\varepsilon=0.025$ of the best predicted viability are in the prescription. The (cumulative) distribution over the cell lines of the maximum $\varepsilon^*$ over the recommended drug for that cell line are shown for a prescription with Dr.S (blue) and a tissue type based prescription (red).}
    \label{fig:DrS_resultfig3}
\end{figure}

We investigate the performance in a more clinically realistic scenario, where we consider a set of Dr.S recommendations that are close enough in predicted viability to our best predicted viability.
Figure \ref{fig:DrS_resultfig3} shows the results for prescribing all drugs that have a predicted viability that is at most $\varepsilon$ higher than the best predicted viability. When Dr.S recommends drugs that have a \emph{predicted} viability within $\varepsilon$ of the best \emph{predicted} viability, the \emph{true} viability of the recommended drugs and their distance to the \emph{true} best viability may be different, and hence the \emph{true} $\varepsilon$, which we denote by $\varepsilon^*$, may be different. Figures \ref{fig:DrS_resultfig3} shows the (cumulative) distribution of $\varepsilon^*$ for each cell line obtained with a prescription $\varepsilon$ equal to 0.025. The distribution corresponding to the Dr.S recommendations clearly lies to the left of the distribution obtained with a tissue type based prescription, and hence Dr.S outperformed a tissue type based prescription.   Dr.S gave a worst case $\varepsilon^*<=0.025$ for 21.4\% 
of the cell lines, 
while this is only 12.3\% 
for the tissue type based prescription. More detailed results can be found in SI Figure 14.

\section{Discussion and conclusions}

The number of FDA approved cancer drugs is growing by about 20 per year and currently exceeds 500 \citep{FDA}. While drugs are generally approved for a specific cancer class, more compounds are coming online which are targeted towards genetic alterations instead of tissue specific cancer types. Other therapies, including radiation and cytotoxic chemotherapies, are more general purpose cell killers, but they too show inter-patient variability that remains largely unexplained. As there are far too many chemical reactions involved to logically piece together what is happening when a drug enters a patient and meets a cancer cell, handling cancer treatment as a big data problem is a promising direction.

In this work we propose the first drug recommendation system for cell lines. Working with cell lines provides access to larger and cleaner datasets and simplifies the model building to cell killing efficacy only, rather than also necessitating building models for side-effects. Given the difficulty of the problem (predicting the effect of a drug on a biological system) we believe it is prudent to start with the simplest representative setting.

Our modeling proceeds in two sequential steps: Model Analysis and Selection (MAS) and Drug Recommendation System (Dr.S). Figure 1, which summarizes the quality of models that are built in MAS, depicts $R^2$ values that seem low (from 0.2 to 0.45), and yet Dr.S does well: if the top four recommendations are considered, the actual best drug is in that batch 50\% of the time (Figure 4b), and about 22\% of the time, the top recommended drug is correct (Figure 4a). This confusion (low $R^2$ but good recommendations) is resolved by considering the calibration plot in Figure 3a: as long as $R^2$ is far enough from 0, the correlation between predicted and true viability is strong enough to obtain a good ranking. Drugs predicted to be the most effective generally are.

In designing the CLA software package, which includes MAS and Dr.S, we put forth the following best practices:
\begin{itemize}
\item \emph{A repeated holdout ``double split'' procedure}: Without access to an independent dataset for final model validation, one usually removes and hides a random set of samples to act as validation data. We repeat this holdout multiple times, which is important when the underlying models are difficult and hence results are highly variable and may depend on the data split. For each outer holdout loop, the inner data is subject to splitting for tuning hyperparameters of the models, hence the use of the phrase ``double split''.
\item \emph{Single drug modeling}: While some authors approach such datasets as the GDSC with a multi-task learning framework \citep{gonen2014drug,menden2013machine,yang2018linking,costello2014community,zhang2018hybrid,cortes2015improved,multitask}, viewing different drugs as different tasks, we instead recommend single drug modeling, since in a clinical setting we will not have the luxury of having each training sample (patient) be treated by all of the drugs we are building models for.
\item \emph{Using prior knowledge for gene selection}: Since our goal is to build a recommendation system rather than discovery of new predictive features, we incorporate prior knowledge in the form of pre-curated gene sets. Because univariate feature selection methods are popular, we demonstrate a statistically valid approach to this, where the strongest univariate features are selected in the inner loop, i.e. not looking at the entire dataset first. Explorations of gene set sizes and random versus curated sets demonstrate that for this dataset, prediction quality generally rises with the number of genes in the sets, at least up to 1500 genes (the maximum we have so far tried) and possibly beyond. For equal size gene sets curated genes outperform randomly selected genes, although the differences in $R^2$ are usually small. 
Curated sets are competitive with large random gene sets in terms of model quality, and superior in terms of interpretability and computation time.
\end{itemize}
Finally, we build a drug recommendation system and report benchmark results for future comparisons (i.e. 22\% of the time our top drug recommendation is correct, Figure 4a, leftmost blue bar). Further improvements will likely come from increased number of samples, but it will be interesting to see if gene set selection, in particular sets customized to the drug being modeled, and where the size of the set is optimized for predictive performance, will have a significant impact.
We expect such new gene sets to be able to improve the performance of Dr.S mostly for those drugs where large random gene sets outperformed the curated gene sets. However, going forward, we recommend customizing the genes used for each drug based on the drug's known biochemical activities, but also assessing large random sets, which may do better.

We take a pan-cancer approach to building the models. It remains to be seen if, given large enough samples of a certain tissue-of-origin, one benefits from building a model using other tissue types as well. In our case, due to the fact that many tissue types were represented by very few cell lines (SI Table 9), we chose to build a single model for all cell types. We also show that the pan-cancer machine learning modeling was superior to the simpler tissue type approach. All of our modeling decisions however are subject to debate: the CLA is designed to explore alternative approaches and is available online. Given the variety of modeling techniques that could be attempted, the community needs fixed standards for comparing techniques. The CLA makes it possible to plug in customized machine learning models which can then be assessed, with data splitting and the generation of summary statistics handled automatically. We have not attempted to be exhaustive in our modeling attempts. We provide a framework and baseline computational results for future comparisons. Several avenues are worth exploring for the improvement of Dr.S:
\begin{itemize}
\item Pathway modeling instead of gene sets: we opted to include prior biological knowledge in the form of gene sets. Pathways, which are essentially gene sets with directed edges, could inform the construction of a neural network \citep{haitham}, or could be used along with the sample specific genomic information to compute sample similarity scores via differential equation modeling \citep{simkern} or the earth-mover's distance \citep{earthmover}. 
\item Other -omics data, such as DNA methylation, chromosomal translocations, and proteomics, may prove to have additional predictive power.
\item In the MAS, we tune models using standard grid search, which could be replaced by Bayesian optimization \citep{bayesianopt} or other advanced strategies. Additionally, we use $R^2$ with the drug-cell line area under the dose response curve as the quality metric, whereas later in Dr.S we judge the system based on ranking. Further explorations might show that using ranking upstream in the MAS might prove a better strategy.
\item Because the number of combinations that need to be explored increases exponentially with the number of gene sets, and in order to precaution against overfitting the entire dataset by simply ``trying too many things'', we limited ourselves to six gene sets. However, it is worth investigating gene sets specialized for the therapy at hand. 
\item Many more machine learning algorithms than the three we consider here are available. Rather than searching for the very best machine learning algorithm available for the drug sensitivity prediction problem, it was our aim to provide a framework that can compare machine learning methods in a fair way and use it to set up Dr.S. As the number of available algorithms is huge, we decided to limit ourselves to three common and trusted methods and leave the testing of other approaches to future research. 
\item It would be ideal clinically to learn an entire dose response curve rather than a single statistic of it, such as AUC, IC50, or viability at a given concentration, but this requires additional modeling, for example to enforce monotonicity of the learned dose response curve.
\item In the present study we focused on recommending a single drug, while in current practice combinations of drugs can be prescribed as well. To allow for the recommendation of combinations of drugs data is required on the combined effect of drugs.
\item Similar to matrix completion algorithms, but applied in a way that is clinically realizable, one can investigate the use of a fixed panel of drugs for which we assume we have the response data for all cell lines (both training and test sets). Responses from the drugs on this panel could be used as \emph{features} for predicting response to drugs not on the panel. Clinically this could be achieved by using a BH3-profiling array, which rapidly tests tumor biopsies against drugs and measures the onset of apoptosis \citep{letai1}.
\end{itemize}


In order to have every patient benefit from genomically-informed medicine, not just the ones who have a mutation that matches an FDA approved drug, we need to build models that predict how patients will respond to drugs, regarding both efficacy and toxicities. To get there, we need large well-curated datasets, such as those being created by efforts like Count Me In \citep{countmein}, biomedical devices to economically capture relevant patient features such as liquid biopsy \citep{liquidbiopsy}, and careful data handling strategies and machine learning algorithms that leverage existing biological knowledge. Here we show how best practices in machine learning may be combined to create a drug recommendation system that outperforms tissue type based recommendations. Such a system will provide a much-needed AI-based guidance for doctors, and contributes to improved, personalized treatment strategies for patients.



\section{Acknowledgements}
MB and AS are supported by the Netherlands Organization for Scientific Research (NWO) Vidi grant 639.072.309. MB and BED are supported by NWO Vidi grant 864.14.004. MB is supported by the ``B-beurs'' from the Catharina van Tussenbroek fund. DC is partially supported by the Therapy Imaging Program (TIP) funded by the Federal Share of program income earned by Massachusetts General Hospital on C06CA059267, Proton Therapy Research and Treatment Center and by RaySearch Laboratories.

\bibliographystyle{plainnat}
\bibliography{refs}

\end{document}